\PassOptionsToPackage{table,x11names}{xcolor}
\documentclass{article}

\usepackage{bookmark}
\PassOptionsToPackage{numbers, compress}{natbib}

    \usepackage[preprint]{neurips_2025}

\usepackage{tcolorbox}
\usepackage{pdflscape}

\usepackage[utf8]{inputenc} 
\usepackage[T1]{fontenc}    
\usepackage{hyperref}       
\usepackage{url}            
\usepackage{booktabs}       
\usepackage{amsfonts}       
\usepackage{nicefrac}       
\usepackage{microtype}      
\usepackage{xcolor}       
\usepackage{tabularx}
\usepackage{graphicx}
\usepackage{subcaption} 
\usepackage{float}      

\hypersetup{
colorlinks= true,
urlcolor = blue,
linkcolor = red,
citecolor = blue,
}

\usepackage{xspace}
\usepackage{graphicx}
\usepackage{multicol}
\usepackage{multirow}
\usepackage{makecell}
\PassOptionsToPackage{numbers}{natbib}
\usepackage{wrapfig}
\usepackage{subcaption}
\usepackage{enumitem}
\usepackage{titletoc}
\usepackage{amsmath}
\usepackage{amssymb}
\usepackage{booktabs}    
\usepackage{longtable}   
\usepackage{array}       
\usepackage{pifont}

\usepackage{listings}
\setlist[itemize]{leftmargin=*}

\title{Automated Generation of Custom MedDRA Queries Using SafeTerm Medical Map}

\author{%
\textbf{Francois Vandenhende$^{1}$ \quad Anna Georgiou$^{1}$ \quad Michalis Georgiou$^{1}$}\\
\textbf{Theodoros Psaras$^{1}$ \quad Ellie Karekla$^{1}$ \quad Elena Hadjicosta$^{1}$} \\[6pt]
$^1$ClinBAY Limited, Limassol, Cyprus\\
Correspondence: \texttt{francois@clinbay.com}\\
\url{https://app.clinbay.com/safeterm}
}

\PassOptionsToPackage{numbers}{natbib}

\begin{document}

\maketitle

\begin{abstract}
\noindent
\textbf{Background:} In pre-market drug safety review, grouping related adverse event terms into standardised MedDRA queries or the FDA Office of New Drugs Custom Medical Queries (OCMQs) is critical for signal detection.

\textbf{Objective:} We present a novel quantitative artificial intelligence system that understands and processes medical terminology and automatically retrieves relevant MedDRA Preferred Terms (PTs) for a given input query, ranking them by a relevance score using multi-criteria statistical methods.

\textbf{Methods:} The system (SafeTerm) embeds medical query terms and MedDRA PTs in a multidimensional vector space, then applies cosine similarity and extreme-value clustering to generate a ranked list of PTs. Validation was conducted against the FDA OCMQ v3.0 (104 queries), restricted to valid MedDRA PTs. Precision, recall and F1 were computed across similarity-thresholds.

\textbf{Results:} High recall (>95\%) is achieved at moderate thresholds. Higher thresholds improve precision (up to 86\%). The optimal threshold (~0.70 - 0.75) yielded recall ~50\% and precision ~33\%. Narrow-term PT subsets performed similarly but required slightly higher similarity thresholds.

\textbf{Conclusions:} The SafeTerm AI-driven system provides a viable supplementary method for automated MedDRA query generation. A similarity threshold of ~0.60 is recommended initially, with increased thresholds for refined term selection.
\end{abstract}


\section{Introduction}
\section{Introduction}

The Medical Dictionary for Regulatory Activities\textsuperscript{\textregistered} (MedDRA\textsuperscript{\textregistered}) is the internationally recognised terminology system for coding adverse event (AE) terms in clinical trials and pharmacovigilance~\cite{ref1}. A crucial safety signal-detection task is grouping semantically related AE terms (which may appear under different preferred terms or hierarchies) so that meaningful adverse-event clusters can be analysed consistently (e.g., ``Hypoglycaemia'' vs.\ ``Blood glucose decreased''). Failure to group related events can dilute the apparent incidence of safety signals.

Traditionally, safety experts use Standardised MedDRA Queries (SMQs)~\cite{ref2} or the FDA Office of New Drugs Custom Medical Queries (OCMQs)~\cite{ref3,ref4} to define clinically meaningful term sets for signal detection. OCMQs are developed by clinical reviewers at the U.S.\ FDA's Office of New Drugs for pre-market safety work, focusing on frequently labelled or important drug-related reactions. For example, the Insomnia OCMQ may group PTs such as \textit{Initial insomnia} and \textit{Early morning awakening} to estimate incidence more reliably. OCMQ~v3.0 contains 104 queries covering terms up to MedDRA~v26.0.

However, the labour required to build and maintain such query sets (especially as MedDRA evolves) is substantial. Advances in natural language processing (NLP), including large-language models (LLMs) combined with embedding methods, offer a route to automated or semi-automated generation of term groupings. Prior studies have shown that embedding models can improve MedDRA term retrieval accuracy~\cite{ref5}, and that cosine-similarity search on embedding spaces can help adverse-drug-event normalization~\cite{ref6}.

Inspired by these developments, we developed an AI-driven pipeline (SafeTerm) that maps an input medical concept query to related MedDRA PTs from a specified version. Our system uses a fine-tuned LLM-based medical corpus encoder, combined with multivariate statistical methods including clustering and extreme-value analysis, to score and rank candidate PTs. In this paper, we describe the method and evaluate its performance against the FDA OCMQ gold standard.
\section{Material and Methods}

\subsection{System Overview -- SafeTerm}

We developed the \textit{SafeTerm Medical Map}, which embeds MedDRA Preferred Terms (PTs) and generic drug names into a unified multidimensional vector space, along with a two-dimensional projection forming a medical map. The core components include a fine-tuned transformer embedding model for medical terminology encoding, and multivariate statistical modules for dimensional reduction (e.g., principal component analysis), clustering (e.g., $k$-means clustering), and canonical correlation analysis for term relationships. 

SafeTerm is already used in pre-market clinical trial and pharmacovigilance contexts for tasks including MedDRA/ATC coding, disproportionality analysis, aggregated signal detection, and query generation. A freely accessible online version is available to registered MedDRA users at:

\begin{center}
\url{https://app.clinbay.com/safeterm}
\end{center}

\subsection{Automated MedDRA Query (AMQ) Pipeline}

\begin{figure}[ht]
		\centering
		\includegraphics[width=0.6\textwidth]{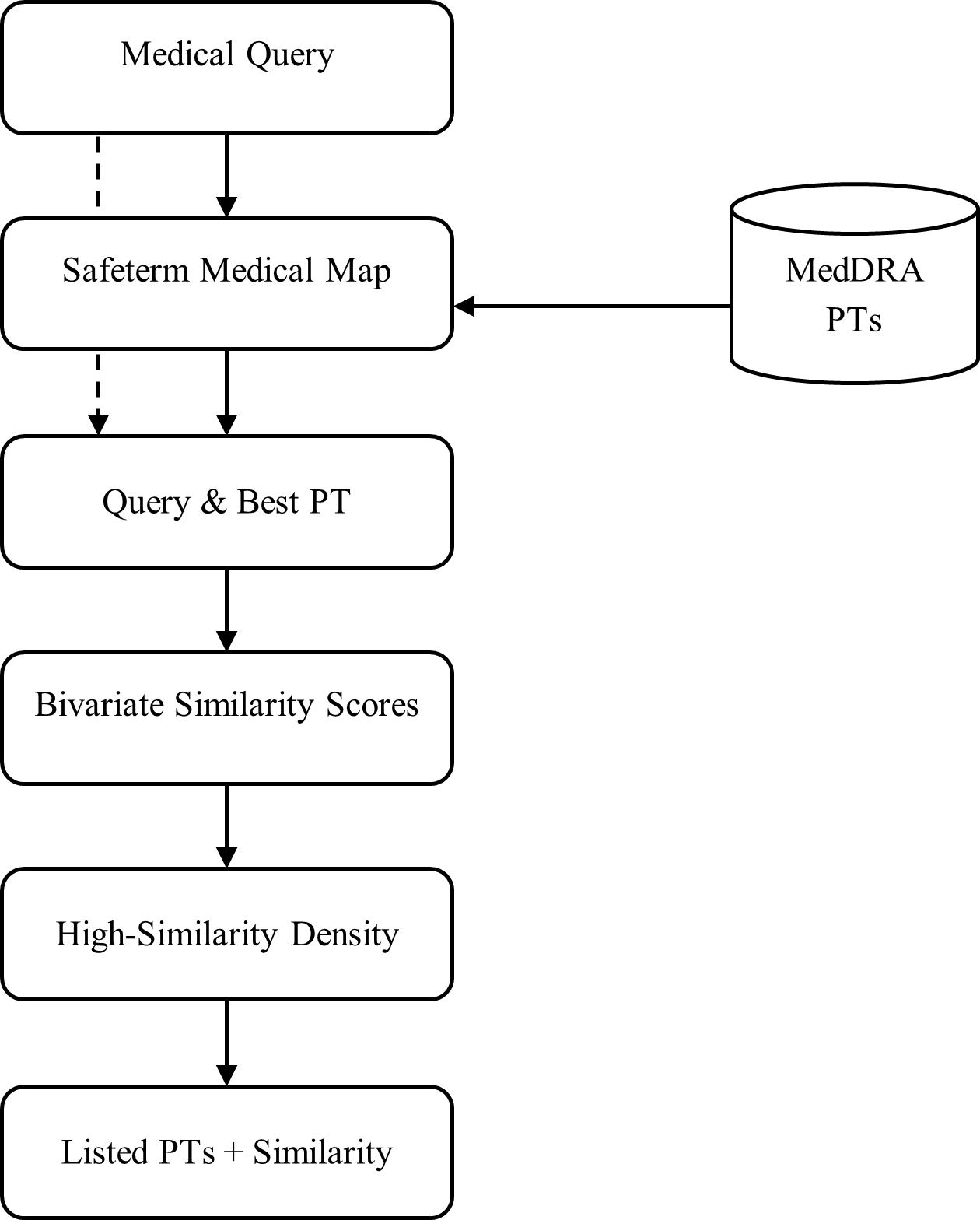}
		\caption{SafeTerm Automated Medical Query System.}
		\label{fig:amq}
	\end{figure}

This application of SafeTerm retrieves a ranked list of MedDRA PTs for any input medical query. The query pipeline (see Figure~\ref{fig:amq}) operates as follows:

\paragraph{Best Term Matching.}
The input query is embedded and projected onto the SafeTerm MedDRA Map to retrieve the most relevant MedDRA Preferred Term (PT). Fuzzy string matching is first applied using a large similarity threshold. If any PT exceeds this threshold, the PT with the highest similarity score is selected. If no syntactic match is found, semantic similarity is assessed using cosine similarity. Up to three PTs with the highest similarity scores are retrieved. If multiple PTs are returned, their embeddings are averaged to form a composite representation.

\paragraph{Similarity Scoring and Clustering.}
Bivariate cosine similarity scores are computed between the query embedding, $\mathrm{sim}(\text{query})$, and the best-matching PT embedding, $\mathrm{sim}(\text{best PT})$, against all MedDRA PTs. A scoring and two-means clustering algorithm is applied to identify the subset of PTs with the highest combined similarity.

\paragraph{Term Ranking.}
The PTs within the retained cluster are then ranked in decreasing order of $\mathrm{sim}(\text{best PT})$.

The AMQ algorithm is deterministic (case-sensitive) and reproducible. It was not fine-tuned on OCMQ/SMQ gold sets, nor does it rely on instructions from the OCMQ/SMQ manuals.

\subsection{Validation against OCMQ v3.0}

We validated the AMQ pipeline against the FDA OCMQ v3.0 reference set of 104 medical-query concepts. Because the OCMQ lists contain term types other than current MedDRA PTs (such as lower-level terms (LLTs), algorithmic terms, or obsolete terms) and draw from multiple MedDRA versions (up to v26.0), we conducted sanitisation so that only valid PTs in the latest MedDRA version~v28.1 were retained.

For each input concept, the AMQ generated a list of candidate PTs, with similarity scores indicating term relevance. We then computed:

\begin{itemize}
    \item \textbf{True Positives (TP)} = PTs present in both OCMQ gold list and AMQ list.
    \item \textbf{Precision} = $\mathrm{TP} \div$ number of PTs in AMQ list.
    \item \textbf{Recall} = $\mathrm{TP} \div$ number of PTs in the OCMQ gold list.
    \item \textbf{F1} = harmonic mean of precision and recall.
\end{itemize}

These metrics were computed at multiple similarity cut-offs from 0.50 to 0.90. We also analysed performance in the subgroup restricted to Narrow PTs only (OCMQ sets labelled Narrow vs.\ Broad). Summary statistics (mean/SD/min/max) were reported across all 104 concepts.
\section{Results}

\subsection{Sanitised OCMQ Set}

Across 85 of the 104 concepts, a total of 692 terms were excluded during sanitisation (median = 4 exclusions per affected concept). The three concepts with the highest exclusions were: \textit{Haemorrhage} ($n=102$), \textit{Malignancy} ($n=78$), and \textit{Diarrhoea} ($n=58$). Most exclusions were due to LLTs or non-PT terms (see Table~1).

\subsection{AMQ Retrieval Performance}

\begin{figure}[ht]
		\centering
		\includegraphics[width=1.0\textwidth]{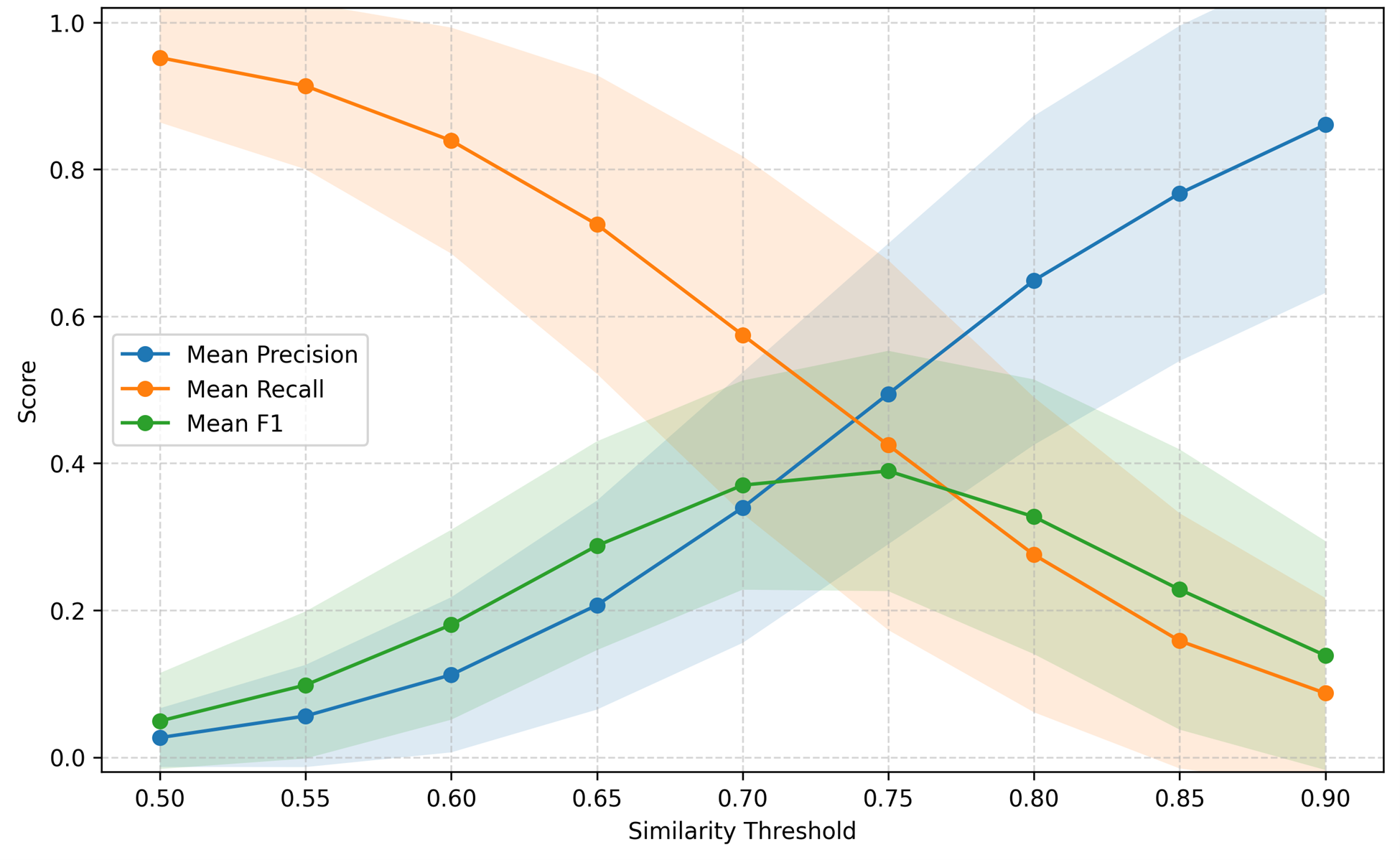}
		\caption{Mean $\pm$ SD performance across similarity cut-offs.}
		\label{fig:cutoff}
	\end{figure}

Table~\ref{tab2} and Figure~\ref{fig:cutoff} summarise retrieval performance (precision, recall, F1) across the 104 OCMQs as a function of similarity cut-off.

Major observations:

\begin{itemize}
    \item At similarity cut-off = 0.50, mean recall $\sim 0.95$, indicating strong sensitivity of the AMQ retriever.
    \item At cut-off = 0.75, mean F1 (0.39; SD 0.16) is maximized, with mean precision $\sim 0.49$ (SD 0.20) and mean recall $\sim 0.42$ (SD 0.25).
    \item At the largest cut-off = 0.90, mean precision $\sim 0.86$ (SD 0.23), showing increasing proportion of valid terms in the AMQs with increasing similarity threshold.
\end{itemize}

As expected, recall decreases and precision increases as the similarity threshold rises. The optimal F1 ($\sim 0.39$) occurred near a threshold of 0.75.

Table~3 summarises the similarity threshold that maximized the F1 score for each OCMQ, together with the corresponding precision, recall, number of true positives, and the sizes of the AMQ-predicted and reference OCMQ term sets. At the optimal similarity threshold (typically 0.70--0.75), and despite wide variation in the number of terms across OCMQs, the number of PTs retrieved by the AMQ procedure was generally comparable to that in the reference lists. Recall exceeded 40\% for most OCMQs, indicating that the automated method captured a substantial proportion of relevant terms. The maximum F1 score was not strongly correlated with the number of terms in the OCMQ reference sets (Figure~3; Pearson $r=0.12$), indicative of broadly similar AMQ performance for both small and large query lists.

\begin{figure}[ht]
		\centering
		\includegraphics[width=1.0\textwidth]{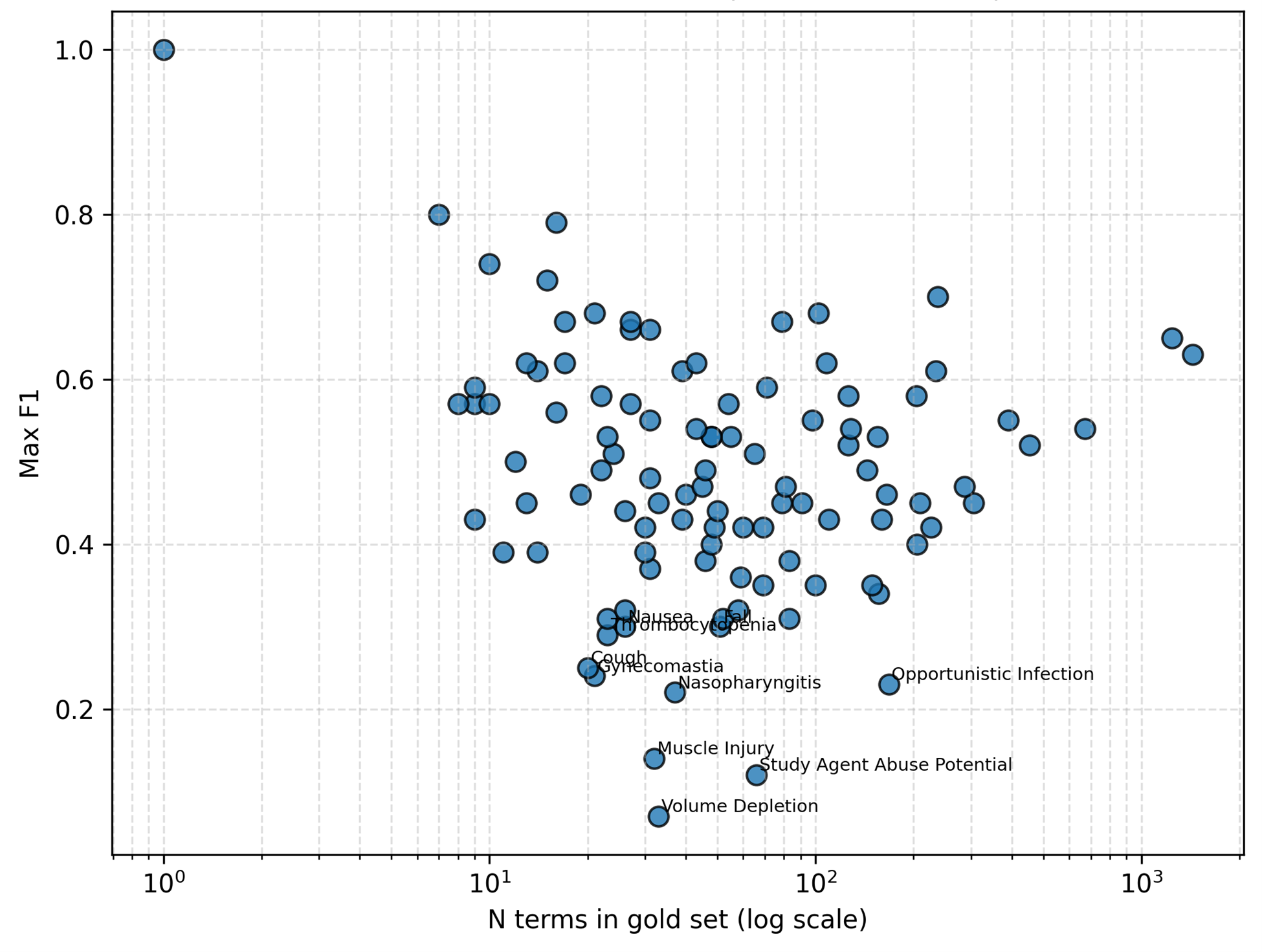}
		\caption{Performance for OCMQ Narrow Terms Retrieval.}
		\label{fig:3}
	\end{figure}

As shown in Table~\ref{tab2} and Figure~\ref{fig:3}, the lowest-performing OCMQs (maximum F1 $<0.2$) included \textit{Study Agent Abuse Potential} (non-MedDRA term), \textit{Muscle Injury} (valid PT), and \textit{Volume Depletion} (defined as an LLT). To emphasise the importance of appropriate query terminology, we replaced the medical-concept LLT \textit{Volume Depletion} with its corresponding PT, \textit{Hypovolaemia}, and re-ran the AMQ request. The maximum F1 score increased from 0.07 to 0.34 (threshold = 0.75 vs.\ 0.60), with corresponding recall and precision of 0.36 (vs.\ 0.18) and 0.32 (vs.\ 0.04), respectively. This improvement highlights the critical role of correct MedDRA terminology in optimising automated query performance.

The AMQ algorithm was not specifically designed to distinguish between narrow and broad PTs. However, higher similarity scores are generally observed for narrow terms. As shown in Figure~\ref{fig:4}, recall increased slightly while precision decreased for narrow-term retrieval, resulting in a modest rightward shift of the F1 curve (by approximately 0.05), with overall magnitude similar to that seen in the combined (broad + narrow) analysis. The optimal F1 was achieved at a similarity threshold between 0.75 and 0.80.

	\begin{figure}[ht]
		\centering
		\includegraphics[width=1.0\textwidth]{figures/fig04.png}
  \caption{Mean $\pm$ SD: Precision, Recall, F1 versus similarity cut-off for Narrow-Term retrieval (across OCMQs).}
		\label{fig:4}
	\end{figure}

\section{Discussion}

We developed and validated an automated MedDRA query (AMQ) retrieval system based on statistical similarity analysis of the SafeTerm MedDRA Map. Unlike traditional approaches, the AMQ algorithm was not fine-tuned or calibrated using SMQs or OCMQs. It operates as a fully independent, unsupervised system that relies solely on syntactic and semantic similarity measures to identify relevant PTs. This independence ensures objectivity and reproducibility across MedDRA versions and use cases.

The system’s performance metrics demonstrate that AMQ is a viable and practical solution for automated term retrieval. Although the overall F1 score ($\sim 0.39$) indicates moderate agreement with FDA-curated OCMQs, this level of accuracy is comparable to inter-expert variability when two human reviewers independently construct query lists. Indeed, several OCMQ reference lists contained non-MedDRA terms or LLTs, which likely limited achievable concordance. In practice, the AMQ’s high recall --- often exceeding 90\% at moderate similarity thresholds --- is much larger than previous attempts~\cite{ref8} and supports its use as a first-pass retrieval tool. Users can then apply expert clinical judgment to refine the list, balancing recall and precision according to their analytic goals.

A key advantage of AMQ lies in its versatility and ease of maintenance across MedDRA versions. Because the system depends only on the query term and the active MedDRA dictionary, PT retrieval automatically adapts to version updates without manual intervention. This approach also allows for query optimization: improving the specificity of the query phrase (e.g., using established PTs or medically precise wording) can directly enhance retrieval accuracy. As a result, AMQ offers an efficient, version-agnostic alternative to manually maintained static term lists.

Although the system does not explicitly differentiate between narrow and broad PTs, the results indicate that similarity scores are typically higher for narrow (more specific) terms, as expected. Performance in the narrow-term subset was comparable to that of the full (broad + narrow) analysis, with the F1 curve shifted rightward by approximately 0.05 in threshold space. This confirms that semantic similarity corresponds to clinical specificity --- supporting the use of higher thresholds (0.75--0.80) when the goal is to identify narrow, high-confidence terms.

\section{Conclusions}

We introduce an automated MedDRA query-generation system using AI and statistical methods. Validated against FDA OCMQ v3.0, it achieves balanced performance, comparable to expert-defined lists, and offers a practical tool for pharmacovigilance professionals. The method is fully unsupervised, reproducible, and adaptable across MedDRA versions, requiring only the input query and current dictionary.

In practice, users should begin with a moderate similarity threshold ($\sim$0.55--0.60) to capture most relevant PTs (recall $\sim$90\%) and review the listed terms by decreasing similarity score --- the most relevant terms appearing on top. Query specificity using proper MedDRA terminology directly improves retrieval accuracy.

\bibliographystyle{unsrtnat}
\bibliography{main}

\clearpage

\begin{table}[ht]
\centering
\caption{Top 3 OCMQs with non-valid MedDRA PT terms excluded}
\label{tab1}
\begin{tabularx}{\textwidth}{l c X}
\hline
\textbf{OCMQ} & \textbf{Number of terms excluded} & \textbf{Sample terms excluded} \\
\hline
Hemorrhage & 102 & Abdominal aortic aneurysm haemorrhage; Anastomotic ulcer haemorrhage, obstructive; Application site bleeding; … \\
Malignancy & 78  & Adenoid cystic carcinoma of external auditory canal; Adrenal carcinoma; Anaplastic large-cell lymphoma; Blastic plasmacytoid; … \\
Diarrhea   & 58  & Antidiarrheal supportive care; Diarrhoea NOS; Diarrhoea aggravated; Loose stools; Stools watery; … \\
\hline
\end{tabularx}
\end{table}
\clearpage

\begin{table}[ht]
\centering
\caption{AMQ Performance Summary across 104 OCMQs}
\label{tab2}
\begin{tabularx}{\textwidth}{c|cccc|cccc|cccc}
\toprule
\textbf{Similarity} & \multicolumn{4}{c|}{\textbf{Precision}} & \multicolumn{4}{c|}{\textbf{Recall}} & \multicolumn{4}{c}{\textbf{F1}} \\
\textbf{Cut-off} & Mean & SD & Min & Max & Mean & SD & Min & Max & Mean & SD & Min & Max \\
\midrule
0.50 & 0.03 & 0.04 & 0.00 & 0.29 & 0.95 & 0.09 & 0.39 & 1.00 & 0.05 & 0.07 & 0.00 & 0.45 \\
0.55 & 0.06 & 0.07 & 0.00 & 0.49 & 0.91 & 0.11 & 0.27 & 1.00 & 0.10 & 0.10 & 0.00 & 0.63 \\
0.60 & 0.11 & 0.11 & 0.00 & 0.65 & 0.84 & 0.15 & 0.18 & 1.00 & 0.18 & 0.13 & 0.01 & 0.65 \\
0.65 & 0.21 & 0.14 & 0.01 & 0.70 & 0.73 & 0.20 & 0.06 & 1.00 & 0.29 & 0.14 & 0.01 & 0.70 \\
0.70 & 0.34 & 0.18 & 0.00 & 0.88 & 0.57 & 0.24 & 0.00 & 1.00 & 0.37 & 0.14 & 0.00 & 0.67 \\
0.75 & 0.49 & 0.20 & 0.00 & 0.91 & 0.42 & 0.25 & 0.00 & 1.00 & 0.39 & 0.16 & 0.00 & 0.74 \\
0.80 & 0.65 & 0.22 & 0.00 & 1.00 & 0.28 & 0.21 & 0.00 & 1.00 & 0.33 & 0.19 & 0.00 & 0.79 \\
0.85 & 0.77 & 0.23 & 0.00 & 1.00 & 0.16 & 0.17 & 0.00 & 1.00 & 0.23 & 0.19 & 0.00 & 1.00 \\
0.90 & 0.86 & 0.23 & 0.00 & 1.00 & 0.09 & 0.13 & 0.00 & 1.00 & 0.14 & 0.16 & 0.00 & 1.00 \\
\bottomrule
\end{tabularx}
\end{table}

\clearpage
\begin{landscape} 
    \setlength{\tabcolsep}{4pt} 
    \renewcommand{\arraystretch}{0.9} 
    \small 
    \begin{longtable}{p{3.5cm}ccccccc} 
    \caption{Per-OCMQ Performance Summary (sorted by descending max F1)} 
    \label{tab3} \\
    \toprule
    \textbf{OCMQ} & \textbf{Similarity Cut-off} & \textbf{Max F1} & \textbf{Precision} & \textbf{Recall} & \textbf{TP (N)} & \textbf{AMQ Prediction (N)} & \textbf{Reference Set (N)} \\
    \midrule
    \endfirsthead
    
    \toprule
    \textbf{OCMQ} & \textbf{Similarity Cut-off} & \textbf{Max F1} & \textbf{Precision} & \textbf{Recall} & \textbf{TP (N)} & \textbf{AMQ Prediction (N)} & \textbf{Reference Set (N)} \\
    \midrule
    \endhead
    
    \midrule
    \multicolumn{8}{r}{\textit{Continued on next page}} \\
    \midrule
    \endfoot
    
    \bottomrule
    \endlastfoot
Palpitations & 0.85 & 1.00 & 1.00 & 1.00 & 1 & 1 & 1 \\
Hyperprolactinemia & 0.85 & 0.80 & 0.75 & 0.86 & 6 & 8 & 7 \\
Tremor & 0.80 & 0.79 & 0.92 & 0.69 & 11 & 12 & 16 \\
Dry Mouth & 0.80 & 0.74 & 0.78 & 0.70 & 7 & 9 & 10 \\
Amenorrhea & 0.80 & 0.72 & 0.90 & 0.60 & 9 & 10 & 15 \\
Fungal Infection & 0.65 & 0.70 & 0.66 & 0.74 & 176 & 267 & 237 \\
Renal \& Urinary Tract Infection & 0.80 & 0.68 & 0.67 & 0.69 & 70 & 104 & 102 \\
Hypoglycemia & 0.75 & 0.68 & 0.65 & 0.71 & 15 & 23 & 21 \\
Self-Harm & 0.70 & 0.67 & 0.69 & 0.65 & 11 & 16 & 17 \\
Seizure & 0.70 & 0.67 & 0.62 & 0.73 & 58 & 93 & 79 \\
Vertigo & 0.75 & 0.67 & 0.67 & 0.67 & 18 & 27 & 27 \\
Diabetic Ketoacidosis & 0.75 & 0.66 & 0.61 & 0.71 & 22 & 36 & 31 \\
Insomnia & 0.75 & 0.66 & 0.55 & 0.81 & 22 & 40 & 27 \\
Bacterial Infection & 0.60 & 0.65 & 0.55 & 0.80 & 991 & 1815 & 1240 \\
Malignancy & 0.55 & 0.63 & 0.49 & 0.86 & 1236 & 2515 & 1434 \\
Fracture & 0.70 & 0.62 & 0.57 & 0.69 & 75 & 132 & 108 \\
Excessive Menstrual Bleeding & 0.80 & 0.62 & 0.62 & 0.62 & 8 & 13 & 13 \\
Pancreatitis & 0.75 & 0.62 & 0.50 & 0.81 & 35 & 70 & 43 \\
Peripheral Edema & 0.85 & 0.62 & 0.67 & 0.59 & 10 & 15 & 17 \\
Stroke and TIA & 0.70 & 0.61 & 0.59 & 0.63 & 148 & 249 & 234 \\
Angioedema & 0.75 & 0.61 & 0.62 & 0.59 & 23 & 37 & 39 \\
Gout & 0.75 & 0.61 & 0.53 & 0.71 & 10 & 19 & 14 \\
Dysgeusia & 0.80 & 0.59 & 0.62 & 0.56 & 5 & 8 & 9 \\
Arrhythmia & 0.80 & 0.59 & 0.74 & 0.49 & 35 & 47 & 71 \\
Psychosis & 0.70 & 0.58 & 0.69 & 0.51 & 64 & 93 & 126 \\
Thrombosis Venous & 0.75 & 0.58 & 0.51 & 0.67 & 137 & 271 & 204 \\
Alopecia & 0.70 & 0.58 & 0.54 & 0.64 & 14 & 26 & 22 \\
Sexual Dysfunction & 0.75 & 0.57 & 0.80 & 0.44 & 24 & 30 & 54 \\
Parasomnia & 0.75 & 0.57 & 0.64 & 0.52 & 14 & 22 & 27 \\
Anaphylactic Reaction & 0.85 & 0.57 & 0.67 & 0.50 & 4 & 6 & 8 \\
Irritability & 0.80 & 0.57 & 0.80 & 0.44 & 4 & 5 & 9 \\
Syncope & 0.75 & 0.57 & 0.44 & 0.80 & 8 & 18 & 10 \\
Tachycardia & 0.85 & 0.56 & 0.56 & 0.56 & 9 & 16 & 16 \\
Abnormal Uterine Bleeding & 0.75 & 0.55 & 0.55 & 0.55 & 17 & 31 & 31 \\
Leukopenia & 0.75 & 0.55 & 0.51 & 0.58 & 57 & 111 & 98 \\
Thrombosis & 0.70 & 0.55 & 0.62 & 0.50 & 194 & 314 & 390 \\
Viral Infection & 0.65 & 0.54 & 0.55 & 0.53 & 355 & 647 & 670 \\
Anxiety & 0.70 & 0.54 & 0.66 & 0.46 & 59 & 90 & 128 \\
Cardiac Conduction Disturbance & 0.80 & 0.54 & 0.55 & 0.53 & 23 & 42 & 43 \\
Systemic Hypertension & 0.80 & 0.53 & 0.56 & 0.50 & 24 & 43 & 48 \\
Pneumonia & 0.70 & 0.53 & 0.51 & 0.55 & 86 & 168 & 155 \\
Tendinopathy & 0.75 & 0.53 & 0.61 & 0.48 & 23 & 38 & 48 \\
Acute Kidney Injury & 0.75 & 0.53 & 0.39 & 0.84 & 46 & 119 & 55 \\
Dizziness & 0.80 & 0.53 & 0.82 & 0.39 & 9 & 11 & 23 \\
Diarrhea & 0.70 & 0.52 & 0.55 & 0.48 & 61 & 110 & 126 \\
Hemorrhage & 0.65 & 0.52 & 0.42 & 0.70 & 319 & 764 & 453 \\
Cholecystitis & 0.75 & 0.51 & 0.52 & 0.51 & 33 & 64 & 65 \\
Hypotension & 0.80 & 0.51 & 0.52 & 0.50 & 12 & 23 & 24 \\
Erectile Dysfunction & 0.85 & 0.50 & 1.00 & 0.33 & 4 & 4 & 12 \\
Osteoporosis & 0.75 & 0.49 & 0.54 & 0.46 & 21 & 39 & 46 \\
Cachexia & 0.75 & 0.49 & 0.48 & 0.50 & 11 & 23 & 22 \\
Heart Failure & 0.75 & 0.49 & 0.48 & 0.51 & 73 & 152 & 144 \\
Back Pain & 0.70 & 0.48 & 0.63 & 0.39 & 12 & 19 & 31 \\
Urticaria & 0.75 & 0.47 & 0.50 & 0.44 & 20 & 40 & 45 \\
Thrombosis Arterial & 0.75 & 0.47 & 0.47 & 0.47 & 136 & 287 & 287 \\
Lipid Disorder & 0.75 & 0.47 & 0.38 & 0.62 & 50 & 132 & 81 \\
Purulent Material & 0.65 & 0.46 & 0.33 & 0.78 & 129 & 390 & 165 \\
Bacterial Vaginosis & 0.75 & 0.46 & 0.33 & 0.74 & 14 & 42 & 19 \\
Urinary Retention & 0.80 & 0.46 & 0.53 & 0.40 & 16 & 30 & 40 \\
Decreased Menstrual Bleeding & 0.75 & 0.45 & 0.56 & 0.38 & 5 & 9 & 13 \\
Glaucoma & 0.70 & 0.45 & 0.44 & 0.45 & 41 & 93 & 91 \\
Anemia & 0.75 & 0.45 & 0.39 & 0.53 & 42 & 107 & 79 \\
Arthritis & 0.60 & 0.45 & 0.34 & 0.64 & 133 & 386 & 209 \\
Hypersensitivity & 0.60 & 0.45 & 0.35 & 0.61 & 187 & 532 & 306 \\
Vomiting & 0.70 & 0.45 & 0.48 & 0.42 & 14 & 29 & 33 \\
Dyspepsia & 0.75 & 0.44 & 0.46 & 0.42 & 11 & 24 & 26 \\
Acute Coronary Syndrome & 0.75 & 0.44 & 0.53 & 0.38 & 19 & 36 & 50 \\
Respiratory Failure & 0.70 & 0.43 & 0.33 & 0.64 & 70 & 213 & 110 \\
Constipation & 0.85 & 0.43 & 0.60 & 0.33 & 3 & 5 & 9 \\
Hyperglycemia & 0.70 & 0.43 & 0.55 & 0.36 & 57 & 103 & 160 \\
Myocardial Infarction & 0.75 & 0.43 & 0.42 & 0.44 & 17 & 40 & 39 \\
Pyrexia & 0.70 & 0.42 & 0.42 & 0.43 & 21 & 50 & 49 \\
Respiratory Depression & 0.70 & 0.42 & 0.37 & 0.48 & 29 & 79 & 60 \\
Fatigue & 0.75 & 0.42 & 0.56 & 0.33 & 10 & 18 & 30 \\
Depression & 0.70 & 0.42 & 0.51 & 0.36 & 25 & 49 & 69 \\
Hepatic Injury & 0.65 & 0.42 & 0.33 & 0.60 & 136 & 415 & 226 \\
Pruritus & 0.75 & 0.40 & 0.64 & 0.29 & 14 & 22 & 48 \\
Local Administration Reaction & 0.60 & 0.40 & 0.26 & 0.89 & 182 & 699 & 205 \\
Confusional State & 0.70 & 0.39 & 0.32 & 0.50 & 15 & 47 & 30 \\
Decreased Appetite & 0.65 & 0.39 & 0.25 & 0.86 & 12 & 48 & 14 \\
Somnolence & 0.75 & 0.39 & 0.27 & 0.73 & 8 & 30 & 11 \\
Myocardial Ischemia & 0.70 & 0.38 & 0.35 & 0.40 & 33 & 93 & 83 \\
Headache & 0.65 & 0.38 & 0.33 & 0.46 & 21 & 64 & 46 \\
Bronchospasm & 0.70 & 0.37 & 0.26 & 0.61 & 19 & 72 & 31 \\
Mania & 0.65 & 0.36 & 0.29 & 0.47 & 28 & 98 & 59 \\
Arthralgia & 0.60 & 0.35 & 0.22 & 0.79 & 118 & 526 & 149 \\
Pneumonitis & 0.75 & 0.35 & 0.46 & 0.28 & 19 & 41 & 69 \\
Paresthesia & 0.65 & 0.35 & 0.29 & 0.43 & 43 & 147 & 100 \\
Rash & 0.70 & 0.34 & 0.36 & 0.33 & 51 & 141 & 156 \\
Abdominal Pain & 0.75 & 0.32 & 0.54 & 0.22 & 13 & 24 & 58 \\
Dyspnea & 0.75 & 0.32 & 0.28 & 0.38 & 10 & 36 & 26 \\
Erythema & 0.70 & 0.31 & 0.29 & 0.33 & 17 & 59 & 52 \\
Hepatic Failure & 0.75 & 0.31 & 0.56 & 0.22 & 18 & 32 & 83 \\
Myalgia & 0.80 & 0.31 & 0.56 & 0.22 & 5 & 9 & 23 \\
Fall & 0.70 & 0.30 & 0.36 & 0.25 & 13 & 36 & 51 \\
Nausea & 0.75 & 0.30 & 0.43 & 0.23 & 6 & 14 & 26 \\
Thrombocytopenia & 0.85 & 0.29 & 0.32 & 0.26 & 6 & 19 & 23 \\
Cough & 0.70 & 0.25 & 0.19 & 0.40 & 8 & 43 & 20 \\
Gynecomastia & 0.70 & 0.24 & 0.17 & 0.38 & 8 & 46 & 21 \\
Opportunistic Infection & 0.65 & 0.23 & 0.16 & 0.38 & 64 & 393 & 168 \\
Nasopharyngitis & 0.75 & 0.22 & 0.20 & 0.24 & 9 & 45 & 37 \\
Muscle Injury & 0.65 & 0.14 & 0.09 & 0.38 & 12 & 134 & 32 \\
Study Agent Abuse Potential & 0.60 & 0.12 & 0.08 & 0.27 & 18 & 222 & 66 \\
Volume Depletion & 0.60 & 0.07 & 0.04 & 0.18 & 6 & 141 & 33 \\

\end{longtable}
\end{landscape}
\clearpage

\end{document}